\documentclass{article}
\usepackage{ijcai25}

\usepackage{times}
\usepackage{xcolor}
\usepackage{soul}
\usepackage{url}
\usepackage[hidelinks]{hyperref}
\usepackage[utf8]{inputenc}
\usepackage[small]{caption}
\usepackage{graphicx}
\usepackage{amsmath}
\usepackage{amsthm}
\usepackage{booktabs}
\usepackage{algorithm}
\usepackage{algorithmic}
\usepackage[switch]{lineno}
\usepackage{xspace}
\usepackage{enumitem}
\usepackage{pifont}
\newcommand{\method}{TSA\xspace}
\newcommand{\stitle}[1]{\vspace*{0.4em}\noindent{\bf #1\/}}
\usepackage{multirow}
\usepackage[normalem]{ulem}
\useunder{\uline}{\ul}{}
\usepackage{setspace}
\usepackage{marvosym}
\usepackage{footmisc}
\newcommand{\squishlist}{
	\begin{list}{$\bullet$}
		{ \setlength{\itemsep}{1pt}
			\setlength{\parsep}{1pt}
			\setlength{\topsep}{2.5pt}
			\setlength{\partopsep}{0.5pt}
			\setlength{\leftmargin}{1em}
			\setlength{\labelwidth}{1em}
			\setlength{\labelsep}{0.6em}
		}
	}
	\newcommand{\squishend}{
	\end{list}
}

\newtheorem{theorem}{Theorem}

\title{Exploiting Text Semantics for Few and Zero Shot Node Classification \\ on Text-attributed Graph}

\author{Yuxiang Wang$^{\dagger}$ \quad Xiao Yan$^{\dagger, \ast}$ \quad Shiyu Jin$^{\dagger}$ \quad Quanqing Xu$^{\ddagger}$ \quad Chuang Hu$^{\dagger}$ \\ Yuanyuan Zhu$^{\dagger}$ \quad Bo Du$^{\dagger}$ \quad
Jia Wu$^{\S}$ \quad Jiawei Jiang$^{\dagger,}$\thanks{Corresponding authors.} 
\affiliations
$^{\dagger}$ School of Computer Science, Wuhan University \\
 $^{\ddagger}$ OceanBase \quad $^{\S}$ School of Computing, Macquarie University
\emails
$^{\dagger}$ \{nai.yxwang,syjin,handc,yyzhu,dubo,jiawei.jiang\}@whu.edu.cn \quad
$^{\dagger}$ yanxiaosunny@gmail.com \\
$^{\ddagger}$ xuquanqing.xqq@oceanbase.com \quad $^{\S}$  jia.wu@mq.edu.au
}

\begin{document}

\maketitle
\begin{abstract}
    Text-attributed graph (TAG) provides a text description for each graph node, and few- and zero-shot node classification on TAGs have many applications in fields such as academia and social networks. Existing work utilizes various graph-based augmentation techniques to train the node and text embeddings, while text-based augmentations are largely unexplored. In this paper, we propose \textbf{T}ext \textbf{S}emantics \textbf{A}ugmentation (\method) to improve accuracy by introducing more text semantic supervision signals. Specifically, we design two augmentation techniques, i.e., \textit{positive semantics matching} and \textit{negative semantics contrast}, to provide more reference texts for each graph node or text description. Positive semantic matching retrieves texts with similar embeddings to match with a graph node. Negative semantic contrast adds a negative prompt to construct a text description with the opposite semantics, which is contrasted with the original node and text. We evaluate \method on 5 datasets and compare with 13 state-of-the-art baselines. The results show that \method consistently outperforms all baselines, and its accuracy improvements over the best-performing baseline are usually over 5\%.
\end{abstract}

\section{Introduction}
Text-attributed graph (TAG) \cite{yan2023comprehensive} is a prevalent type of graph-structured data, where each node is associated with a text description. For instance, in a citation network, the papers (i.e., nodes) are linked by the citation relations (i.e., edges), and the abstract of each paper serves as the text description. Few-shot and zero-shot node classification on TAGs (FZNC-TAG) predict the categories of the nodes using a few or even no labeled data since labeled data are expensive to obtain ~\cite{liu2021relative,liu2022few}. The two tasks have many applications in areas such as recommender system~\cite{gao2022graph}, social network analysis~\cite{yu2020enhancing}, and anomaly detection~\cite{noble2003graph}. 

\begin{figure}[t]
    \centering
    \includegraphics[scale=0.55]{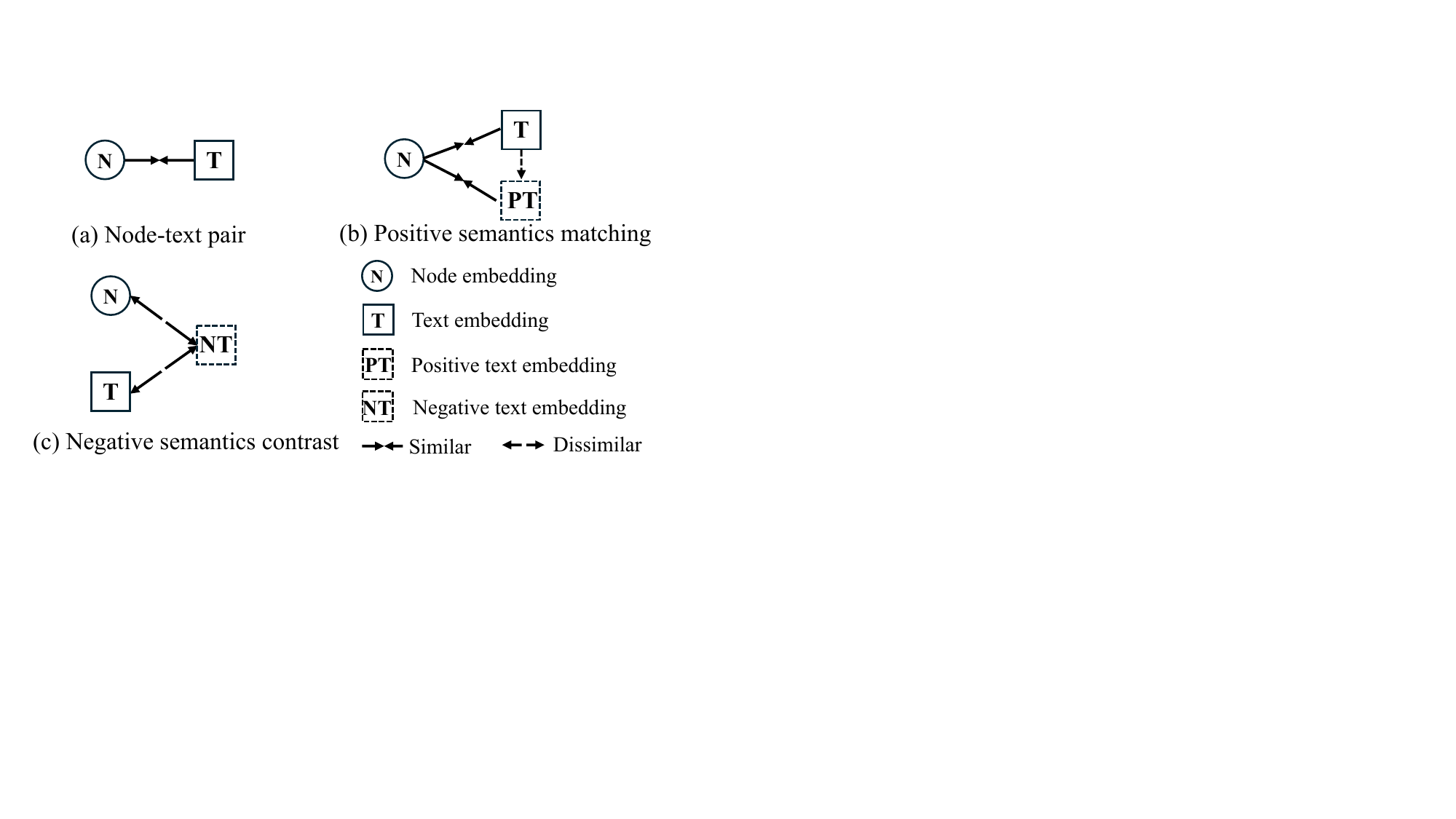}
    \caption{The contrastive loss of G2P2 (a) and two semantic augmentation techniques proposed by \method (b,c). The node-text pair of G2P2 is specified by data as each node has a text description, and \method mines more semantic information for nodes and texts.}
    \label{fig:alpha}
    \vspace{-0.4cm}
\end{figure}

Existing methods for FZNC-TAG typically follow a two-step process: first learn node and text embeddings on the TAGs and then use prompting to produce classification results~\cite{wen2023augmenting,huang2023prompt}. They mainly differ in embedding learning and can be classified into three categories.
\ding{182} Self-supervised graph learning methods employ graph augmentation techniques, such as node adding and feature masking, to generate more generalized node embeddings \cite{you2020graph,deng2021graph}.
They exploit the graph topology but ignores the information in the text semantics.
\ding{183} The two-stage learning methods encode the text using large language models (LLMs) and add the text embedding as additional node features \cite{tang2024graphgpt,chen2024exploring}. Then, graph neural network (GNN) methods, e.g., TextGCN~\cite{yao2019graph} and GraphSAGE~\cite{hamilton2017inductive}, are used to learn the node embeddings. The limitation of these approaches is that the LLMs are not updated during GNN training~\cite{yan2023comprehensive}.
\ding{184} The state-of-the-art end-to-end learning method, G2P2~\cite{wen2023augmenting}, jointly trains the GNN and language model via contrastive learning ~\cite{he2020momentum}. As shown in Figure~\ref{fig:alpha}(a), G2P2 contrasts each node-text pair to ensure that the GNN and language model are aligned in the embedding space.

Existing work utilizes various graph-based augmentation to train the node and text embeddings, however, neglecting the semantics information in the texts. As shown in Figure~\ref{fig:alpha}(a), G2P2 merely aligns the raw node-text pairs in the TAGs. Furthermore, we observe that the classification accuracy of G2P2 is low. For instance, on the Fitness dataset~\cite{yan2023comprehensive}, G2P2 only achieves an accuracy of 68.24\% and 45.99\% for few- and zero-shot classification, respectively. The significantly lower accuracy in zero-shot classification also suggests that the problems are caused by the lack of text semantics and that G2P2’s sole reliance on the graph-based augmentation is inadequate for training high-quality models. Thus, we ask the following research question:
\begingroup
\addtolength\leftmargin{0.001in}
\addtolength\rightmargin{0.001in}
\begin{quote}
    \textit{How to exploit more text semantics to enhance few- and zero-shot classification on TAGs?}
\end{quote}
\endgroup

To answer the question, we present \method, where embedding learning can be improved by enforcing similarity relations among embeddings.  
Inspired by this idea, we design two augmentation techniques, i.e., positive semantics matching and negative semantics contrast, as shown in Figure~\ref{fig:alpha}(b,c). These techniques create additional node-text pairs that have similar or dissimilar embeddings to facilitate model learning.

\stitle{Positive semantics matching.} In Figure \ref{fig:alpha}(b), we provide multiple positive text embeddings for each node embedding. This is achieved by searching the texts that have similar embeddings to the text of the considered graph node. We also encourage the node embedding to be similar to those text embeddings. This augmentation provides more supervision to GNN training and enforces the prior that nodes may belong to the same categories if their texts have similar semantics.

\stitle{Negative semantics contrast.} In Figure \ref{fig:alpha}(c), we pair each text with a semantically opposite \textit{negative text}, which is constructed by adding a learnable negative prompt to the original text. 
We then encourage the original text embeddings and its corresponding node to be dissimilar to the negative text. This augmentation provides additional text semantics to make the classification robust. 
For instance, to classify a paper as being related to artificial intelligence, it should not only be similar to the description \textit{``a paper is published at IJCAI''} but also be dissimilar to \textit{``a paper is published at The Lancet''}.

We conduct extensive experiments to evaluate \method, using 5 datasets and comparing with 13 state-of-the-art baselines. The results show that \method consistently achieves higher accuracy than all baselines for both  few-shot and zero-shot setting. In particular, \method improves the accuracy and F1 scores of few-shot classification by 4.6\% and 6.9\% on average, and zero-shot classification by 8.8\% and 9.3\%, respectively. 

In summary, we make the following contributions:
\begin{itemize}[leftmargin=*]
    \item We observe that prior methods only contrast node-text pairs, thus failing to effectively capture text semantics. To solve this issue, we propose \method to enhance the semantic understanding in both model pre-training and inference.
    \item We incorporate two novel augmentation techniques into \method: positive semantics matching and negative semantics contrast. These techniques create additional node-text pairs by mining more text semantics from diverse perspectives.
    \item We conduct extensive experiments to evaluate \method~and compare with state-of-the-art baselines, demonstrating that \method~enjoys high model accuracy and training efficiency.
\end{itemize}

\section{Preliminaries}
\stitle{Text-attributed graph.} We denote a text-attributed graph (TAG) as $\mathbf{G}=(\mathcal{V},\mathcal{E},\mathbf{X})$, in which $\mathcal{V}$, $\mathcal{E}$, and $\mathbf{X}$ are the node set, edge set, and text set, respectively. Take citation network as an example for TAG. Each node $v_i\in \mathcal{V}$ is a paper, interconnected by the edges $e\in \mathcal{E}$ that signify citation relations. Let $\mathbf{x}_{i}\in \mathbf{X}$ denote the text (i.e., paper abstract) of the $i$-th node. Each node has a label to indicate the topic of the paper. Since the graph nodes and papers have a strict one-to-one correspondence, node $v_{i}$ and text $\mathbf{x}_i$ share an identical label.

\stitle{Few- and zero-shot learning.} For few-shot classification, the test dataset encompasses a support set $\mathcal{S}$ and a query set $\mathcal{Q}$. $\mathcal{S}$ comprises $C$ classes of nodes, with $K$ labeled nodes drawn from each class. These nodes can be used to train or fine-tune the classifier, which is then utilized to classify the nodes in $\mathcal{Q}$. Zero-shot node classification is essentially a special case of few-shot classification with $K=0$. There are no labeled nodes for both  training and testing, and classification depends solely on the class names.

\stitle{Contrastive loss.} Recent researches \cite{wen2023augmenting,zhao2024pre} use the contrastive loss to jointly train the graph and text encoders. Specifically, they employ GCN~\cite{kipf2016semi} as the graph encoder $\phi$ to encode each node $v_i$ into a node embedding $\mathbf{n}_i$, and adopt Transformer~\cite{vaswani2017attention} as the text encoder $\psi$ to map each text $\mathbf{x}_i$ to  a text embedding $\mathbf{t}_i$. That is,
\begin{equation}
    \mathbf{n}_i=\phi (v_i), \quad \mathbf{t}_{i}=\psi (\mathbf{x}_i).
\end{equation}

\begin{figure*}[t]
    \centering
    \includegraphics[scale=0.6]{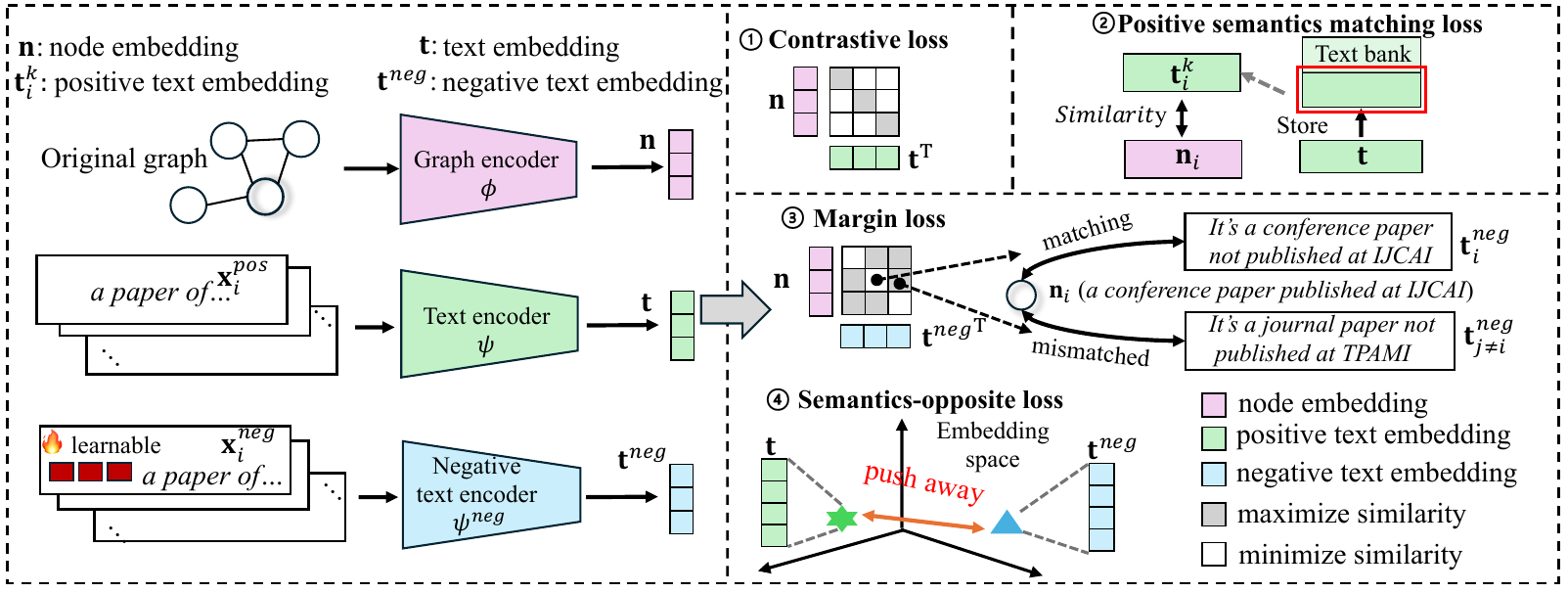}
    \caption{The overview of \method.}
    \label{fig:framework overview}
    \vspace{-0.2cm}
\end{figure*}

Then, they use InfoNCE~\cite{he2020momentum} loss $\mathcal{L}_{CL}$ to maximize the similarity between each node $\mathbf{n}_{i}$ and its corresponding text $\mathbf{t}_{i}$, while simultaneously minimizing the similarity between node $\mathbf{n}_{i}$ and other mismatched texts $\mathbf{t}_{j}$.  As shown in part (1)  of Figure~\ref{fig:framework overview}, $\mathcal{L}_{CL}$ is calculated as follows:
\begin{equation}
    \mathcal{L}_{CL}= -\frac{1}{\left | \mathcal{B} \right | } \!\sum_{(\mathbf{n}_{i},\mathbf{t}_{i})\in \mathcal{B}}^{} \!\mathrm{log} \frac{\mathrm{exp}(\mathrm{sim}(\mathbf{n}_{i}, \mathbf{t}_{i})/\tau) } { {\textstyle \sum_{j\ne i }}  \mathrm{exp}(\mathrm{sim}(\mathbf{n}_{i},\mathbf{t}_{j})/\tau) },
\label{eq:cl loss}
\end{equation}
where $\mathcal{B}$ is a data batch, $\mathrm{sim}(,)$ is the cosine similarity, and $\tau$ is a learnable temperature.

\section{The \method Framework}
In this section, we present a novel pre-training and inference framework, named \method. We start with a overview and follow up with the detailed descriptions of its components.

\subsection{Overview}
The overall architecture of our framework is illustrated in Figure~\ref{fig:framework overview}. The model for few-shot consists of a graph encoder and a text encoder, and an extra negative text encoder is included for zero-shot pre-training. We introduce them as follows.
\begin{itemize}[leftmargin=*]
    \item \textbf{Graph encoder} $\phi$. We adopt a graph neural network as the encoder to generate the node embedding $\mathbf{n}$.
    \item \textbf{Text encoder} $\psi$. We choose Transformer~\cite{vaswani2017attention} as the text encoder, and it produces a text embedding $\mathbf{t}$ for each text description.
    \item \textbf{Negative text encoder} $\psi^{neg}$. This maintains the same architecture as the text encoder, with the difference that we train it independently with a negative prompt to generate the negative text presentation $\mathbf{t}^{neg}$.
\end{itemize}

To effectively train the above encoders, we design two novel loss functions: \textit{positive semantics matching loss} and \textit{negative semantics contrast loss}, which can assist the pre-training model in mining more semantic information.
Next, we propose a strategy in Section~\ref{sec:inference strategy}, \textit{probability-average}, to enhance zero-shot node classification.

\subsection{Text Semantics Augmentation}
\label{sec: ssl pretrain model}
In this section, we introduce two novel augmentations: negative semantics contrast and positive semantics matching, to exploit text semantics to enhance model training on TAGs.

\stitle{Positive semantics matching loss.} We provide more positive text embeddings for each node embedding. G2P2~\cite{wen2023augmenting} defaults to only one text embedding similar to each node embedding, however there may be multiple similar texts to the target node in TAGs~\cite{he2020momentum,chuang2020debiased}. Therefore, we search for multiple text embeddings that are similar to the text embedding of the target node and subsequently encourage the target node embedding to align with these similar text embeddings $\hat{\mathbf{t}}$. The positive semantics matching loss is denoted as follow:
\begin{equation}
    \mathcal{L}_{PSM}=-\frac{1}{\left | \mathcal{B} \right | }\!\sum_{(\mathbf{n}_{i},\mathbf{t}_{i})\in \mathcal{B}}\!\mathrm{log} \frac{ {\textstyle \sum_{k=1}^{\mathbf{K}}} \mathrm{exp}(\mathrm{sim}(\mathbf{n}_i ,\hat{\mathbf{t}}_{i}^{k} )/\tau) } { {\textstyle \sum_{j\ne i}}  \mathrm{exp}(\mathrm{sim}(\mathbf{n}_{i},\mathbf{t}_{j})/\tau) },
    \label{eq:text matching loss}
\end{equation}
where $\mathbf{K}$ is the number of similar text embeddings.

However, the above method has two serious drawbacks: first, the complexity of brute-force search for similar text embeddings among all text embeddings is unacceptable; second, storing all text embeddings in GPU memory may lead to out-of-memory. To address these issues, we create a text bank with a capacity of 32K to model the whole text embedding space. As illustrated in the Figure~\ref{fig:framework overview}(2), whenever a new batch of data arrives, the earliest text embedding is discarded if the capacity of the text bank exceeds a predetermined threshold. Otherwise, it is stored in the bank. Subsequently, we identify the most $\mathbf{K}$ similar text embeddings for target node through similarity calculations. In this way, our text bank is both time-efficient and space-efficient.

\stitle{Negative prompt.} After co-training the graph and text encoder using Equations~\eqref{eq:cl loss} and \eqref{eq:text matching loss}, the model now possesses the base capability to distinguish node-text pairs. However, understanding the negative semantics within the input text description poses a challenge for the model. For example, we represent a text description such as \textit{``a paper is published at IJCAI''} and its negation \textit{``a paper is published at The Lancet''}. In the embedding space, these two descriptions are likely to be very similar, as their raw texts differ by only few words. To address this issue, we employ negative prompts to generate multiple negative texts that are semantically opposed to the original text descriptions. These negative texts are then used to train a negative text encoder independently. This process helps the negative text encoder learn parameters that are contrary to those of the text encoder.

Our initial idea is to manually construct a series of negative prompts. Specifically, we manually alter the text descriptions by incorporating negation terms such as \textit{``no'', ``not'', ``without''}, etc., thus creating a negative prompt corpus that are semantically opposite to the original ones, denoted as $\mathbf{X}^{neg}$. Then, we input the negative text $\mathbf{x}_{i}^{neg}$ into negative text encoder $\psi^{neg}$ to generate negative text embedding $\mathbf{t}^{neg}_i$.

However, manual modification of the raw text is time-consuming and labor-intensive. To solve this problem, inspired by CoOp~\cite{zhou2022conditional}, we propose a \textit{learnable negative prompt} and add it to the front of raw text. The underlying logic is to represent negative semantics by constantly optimizing the learnable prompt, thereby mirroring the hand-crafted negative prompts $\mathbf{X}^{neg}$. Specifically, we concatenate the raw text with $\mathrm{M}$ learnable vectors to generate the negative prompt $\mathbf{h}$, allowing $\mathbf{X}^{neg}$ to be replaced by $\mathbf{h}$. Then we input $\mathbf{h}$ into the negative text encoder $\psi^{neg}$, denoted as follows:
\begin{equation}
    \mathbf{h}=[\underbrace{\mathrm{V}_1, \mathrm{V}_2,... \mathrm{V}_{\mathrm{M} },}_{\mathrm{negative\ prompt}}\mathbf{x} ], \quad \mathbf{t}^{neg}_i = \psi^{neg}(\mathbf{h}_i),
    \label{eq:learnable negative prompt}
\end{equation}
where the negative text encoder is a Transformer with the same architecture as the text encoder. We further demonstrate from the perspective of information entropy that negative text can replace hand-crafted negative prompt.

\begin{theorem}
    
Given the learnable negative prompt $\mathbf{h}$ and hand-crafted negative prompt $\mathbf{X}^{neg}$, the information entropy of the learnable negative prompt $H(\mathbf{h})$ is related to the lower bound of the information entropy of the hand-crafted negative prompt ${\tt LowerBound}(H(\mathbf{X}^{neg}))$ as follow:
\begin{equation}
    H(\mathbf{h})\ge {\tt LowerBound}(H(\mathbf{X}^{neg})).
\end{equation}
\end{theorem}
\noindent This indicates that learnable negative prompts exhibit higher information entropy than the lower bound of hand-crafted negative prompts. Therefore, learnable negative prompts can effectively capture the negative semantics present in hand-crafted negative prompts. Proof see in Appendix A.

\stitle{Negative semantics contrast loss.} There is still an unsolved problem: \textit{how do we train a negative text encoder?} In other words, how do we ensure that the semantics of the negative text embeddings contradict the original text embeddings. To address this problem, we introduce a novel loss functions, termed negative semantics contrast loss, which comprises margin loss and semantics-opposite loss.

The margin loss anticipates the greatest possible similarity between positive pairs, and conversely, it expects dissimilarity in the case of negative pairs.
As shown in Figure~\ref{fig:framework overview}(3), given a target node $v_i$, the corresponding negative text description $\mathbf{t}^{neg}_i$ is deemed a negative text, while any other non-corresponding text $\mathbf{t}^{neg}_{j\ne i}$ are considered positive texts. Subsequently, we employ margin loss to assess the degree of matching between the target nodes, positive texts, and negative texts. Specifically, margin loss ensures that the similarity between the target node and the positive text is at least a margin $m$ higher than the similarity with the negative text. Here, we use margin loss instead of InfoNCE loss because margin loss remains constant when the gap between positive and negative samples exceeds $m$. This is favorable because positive and negative samples are already easily distinguishable. The margin loss $\mathcal{L}_{ML}$ is denoted as follows:
\begin{equation}
    \mathcal{L}_{ML}= max(0, 1+\text{sim}(\mathbf{n}_i,\mathbf{t}^{neg}_{i})-\text{sim}(\mathbf{n}_i,\mathbf{t}^{neg}_{j\ne i} ))
    \label{eq:ml loss}
\end{equation}

As shown in Figure~\ref{fig:framework overview}(4), semantics-opposite loss seeks to maximize the mean square error between positive and negative text embeddings. As text $\mathbf{x}_i$ and negative text $\mathbf{t}^{neg}_i$ are semantically opposite, their corresponding embeddings should be as far apart as possible in the text embedding space. We compute the semantics-opposite loss $\mathcal{L}_{SO}$ as follow:
\begin{equation}
    \mathcal{L}_{SO}=-\frac{1}{\left | \mathcal{B} \right | }\sum_{\mathbf{t}_{i}\in \mathcal{B}}\left \| \mathbf{t}_{i}- \mathbf{t}_{i}^{neg} \right \|_{2},
\end{equation}
where $\left \|  \right \| _2$ is the L2 norm. Thus, the negative semantics contrast loss is equal to the sum of margin loss and semantics-opposite loss, denoted as $\mathcal{L}_{NSC}=\mathcal{L}_{ML}+\mathcal{L}_{SO}$. It enforces both the node and text embeddings are dissimilar to the corresponding negative text embedding.

\stitle{Objective.} In summary, we denote the total loss of \method as:
\begin{equation}
    \mathcal{L}= \mathcal{L}_{CL}+ \mathcal{L}_{PSM} + \alpha \mathcal{L}_{NSC},
\end{equation}
where $\alpha$ is the hyperparameter with respect to loss activation. 
In few-shot pre-training, we do not activate the loss $\mathcal{L}_{NSC}$ (i.e., $\alpha=0$) because the prompts in few-shot are inherently learnable, incorporating negative prompts would introduce more noise and lead to sub-optimal performance. In contrast, zero-shot classification lacks labeled data during the pre-training. 
We analyze the ablation experiments on the loss function in detail in Section~\ref{sec:ablation study}. 

\stitle{Complexity Analysis.} \method incorporates both a GNN and a Transformer. The GNN takes $O(LNd^2)$ time for aggregating the neighboring nodes, where $L$ is the network depth, $N$ is the number of nodes and $d$ is the number of dimensions. The Transformer's time complexity is $O(sd^2+s^2d)$, where $s$ the maximum length of the input sequence. $O(sd^2)$ time is used for mapping vectors at each position to query, key and value vectors, and $O(s^2d)$ time is utilized for the computation of the attention score. Consequently, the overall time complexity of our method is $O(LNd^2+sd^2+s^2d)$.

\subsection{Prompt Tuning and Inference}
\label{sec:inference strategy}
Based on the pre-trained model, we tune the model parameters to adapt to the few/zero-shot tasks, which enables classification with a few even no labeled samples while concurrently freezing the pre-trained model's parameters. 
Prompt tuning is a prompt optimization technique in multimodal models for improving accuracy in few/zero-shot tasks. It uses learnable context vectors $[v]$ to replace the static prompt words. For example, ``a photo of a [class]'' is replaced by $[v_1][v_2][v_3] [class]$.
Next, we introduce the foundational paradigm of few- and zero-shot node classification.

\stitle{Zero-shot classification.} In the zero-shot setting, we operate without any labeled samples and rely solely on class name description. To perform $C$-way node classification, we construct a series of class descriptions $\{\mathbf{D}_c\}_{c=1}^{C}$, via discrete prompts, such as ``\textit{a paper of} [\textit{class}]''. Then, we input the description text into the pre-trained text encoder to generate the class embedding $\mathbf{g}_c=\psi (\mathbf{D}_c)$.
We predict the category of a node $v_i$ by computing the similarity between the node embedding $\mathbf{n}_i$ with the class embedding $\mathbf{g}_c$. The insight behind this is that we align the pre-training and prompting objectives (i.e., to determine whether nodes and texts are similar). Thus, we do not have to tune the parameters of the pre-trained model. The similarity probability between the target node and the candidate class description is calculated as follows:
\begin{equation}
    p_{i}=\frac{\text{exp}(\text{sim}(\mathbf{n}_i,\mathbf{g}_c) /\tau) }{ {\textstyle \sum_{c=1}^{C}\text{exp}(\text{sim}(\mathbf{n}_i,\mathbf{g}_c)/\tau ) } }.
    \label{eq:positive classification probability}
\end{equation}

\begin{figure}[t]
    \centering
    \includegraphics[scale=0.5]{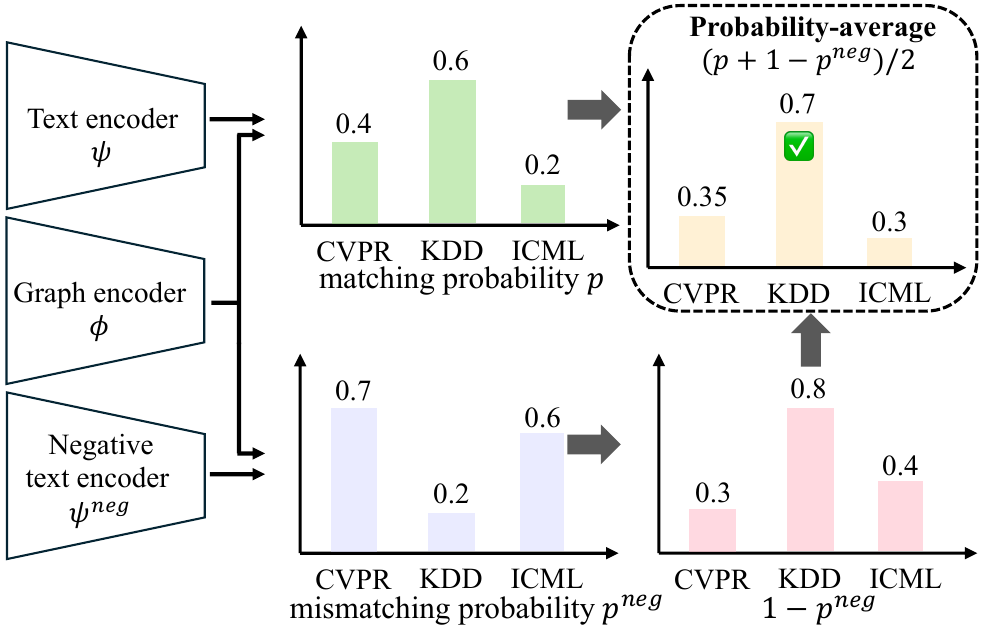}
    \vspace{-0.2cm}
    \caption{The illustration of probability-average.}
    \label{fig:inference stratery}
\end{figure}

\stitle{Few-shot classification.} In the few-shot setting, we conduct a $C$-way $K$-shot classification task. Unlike discrete prompts (i.e., ``\textit{a paper of ...}'') in the zero-shot setting, we have $C\times K$ labeled samples to train learnable prompts.
Specifically, we construct a continuous prompt $\mathbf{g}_c$ by adding $M$ learnable vectors to the front of the class description $\mathbf{D}_c$. Formally, we denote $\mathbf{g}_c=\psi ([\mathbf{e}_1, \mathbf{e}_2,...,\mathbf{e}_M,\mathbf{D}_c ] )$. Then, we use Equation~\eqref{eq:positive classification probability} to predict the node category, and update the continuous prompts by minimizing the discrepancy between the predicted and ground-truth labels via cross-entropy loss. It is worth noting that because $C\times K$ is a small value, the parameters required to fine-tune the prompts are considerably less than those needed for the pre-trained model. 

\stitle{Probability-average.} As shown in Figure~\ref{fig:inference stratery}, we propose probability-average to predict node category. Specifically, we first compute $p_{i}$ by Equation ~\eqref{eq:positive classification probability}. We use the negative text encoder to generate the negative class embedding. Then, we compute negative probability $p^{neg}_{i}$ by contrasting these negative class embeddings with the target node embedding using Equation~\eqref{eq:positive classification probability}. $p_{i}$ denotes the probability that a node belongs to each category and vice versa, $p^{neg}_{i}$ represents the probability that a node does not belong to each category. Finally, we utilize $(p_{i}+1-p^{neg}_{i})/2$ to predict the node label. Unlike using a single-encoder model, the negative text encoder provides additional predictive auxiliary for the positive text encoder. Therefore, we use probability-average to balance the output probabilities of the positive and negative text encoders and thus enhancing classification accuracy. Formally, the probability-average strategy can be denoted as follows:
\begin{equation}
    \mathcal{Y}_i =\text{arg} \ \text{max}\ (p_i +1-p_i^{neg})/2.
    \label{eq:probability-average}
\end{equation}

Note that the probability-average stragety is only applicable to zero-shot classification, as it requires the negative prompts and negative text encoder to calculate $p^{neg}_{i}$. In contrast, few-shot classification directly uses $p_{i}$ to predict labels.

\section{Experimental Evaluation}
In this section, we conduct extensive experiments to evaluate \method and answer the following research questions.
\squishlist
\item \textbf{RQ1}: How does \method compare with state-of-the-art methods for few- and zero-shot classification on TAGs? 
\item \textbf{RQ2}: Do our augmentation techniques improve accuracy? 
\item \textbf{RQ3}: How efficient is \method in training and inference?
\squishend

\subsection{Experiment Settings}
\stitle{Datasets.} Following related researches~\cite{yan2023comprehensive}, we use 5 datasets for experiments. Cora~\cite{mccallum2000automating} is a citation network, where papers are linked by citation relations and abstract serves as the text. Art, Industrial, M.I., and Fitness are derived from Amazon product categories~\cite{yan2023comprehensive}, namely, arts, crafts and sewing for Art; industrial and scientific for Industrial; musical instruments for M.I.; and sports-fitness for Fitness, respectively. For the four datasets, an edge is added to construct the graph if a user visits two products successively, and the text is the product description. The five datasets cover different scales (from thousands to millions of nodes) and number of classes (from tens to thousands). The dataset statistics is provided in Appendix B.

\begin{table*}[t]
\centering
\resizebox{\textwidth}{!}{
\begin{tabular}{l|cc|cc|cc|cc|cc}
\toprule
\multirow{2}{*}{\textbf{Method}} & \multicolumn{2}{c|}{\textbf{Cora}} & \multicolumn{2}{c|}{\textbf{Fitness}}         & \multicolumn{2}{c|}{\textbf{M.I.}}         & \multicolumn{2}{c|}{\textbf{Industrial}}   & \multicolumn{2}{c}{\textbf{Art}}          \\ \cline{2-11}
                                 & ACC                 & F1                  & ACC                  & F1                   & ACC                 & F1                  & ACC                 & F1                  & ACC                 & F1                  \\
\midrule
GCN                              & 41.15±2.41          & 34.50±2.23          & 21.64±1.34           & 12.31±1.18           & 22.54±0.82          & 16.26±0.72          & 21.08±0.45          & 15.23±0.29          & 22.47±1.78          & 15.45±1.14          \\
SAGEsup                          & 41.42±2.90          & 35.14±2.14          & 23.92±0.55           & 13.66±0.94           & 22.14±0.80          & 16.69±0.62          & 20.74±0.91          & 15.31±0.37          & 22.60±0.56          & 16.01±0.28          \\
TextGCN                          & 59.78±1.88          & 55.85±1.50          & 41.49±0.63           & 35.09±0.67           & 46.26±0.91          & 38.75±0.78          & 53.60±0.70          & 45.97±0.49          & 43.47±1.02          & 32.20±1.30          \\
\midrule
GraphCL                          & 76.23±1.80          & 72.23±1.17          & 48.40±0.65           & 41.86±0.89           & 67.97±2.49          & 59.89±2.51          & 62.13±0.65          & 54.47±0.67          & 65.15±1.37          & 52.79±0.83          \\
InfoGCL                              & 78.53±1.12          & 74.58±1.24          & 47.56±0.59           & 41.98±0.77           & 68.06±0.73          & 60.64±0.61          & 52.29±0.66          & 45.26±0.51          & 65.41±0.86          & 53.57±0.75          \\
PGCL                         & 76.32±1.25          & 73.47±1.53          & 48.90±0.80           & 41.31±0.71           & 76.70±0.48          & 70.87±0.59          & 71.87±0.61          & 65.09±0.47          & 76.13±0.94          & 65.25±0.31          \\
\midrule
GPPT                             & 75.25±1.66          & 71.16.±1.13         & 50.68±0.95           & 44.13±1.36           & 71.21±0.78          & 54.73±0.62          & 75.05±0.36          & 69.59±0.88          & 75.85±1.21          & 65.12±0.83          \\
GFP                              & 75.33±1.17          & 70.78±1.62          & 48.61±1.03           & 42.13±1.53           & 70.26±0.75          & 54.67±0.64          & 74.76±0.37          & 68.55±0.29          & 73.60±0.83          & 63.05±1.61          \\
GraphPrompt                      & 76.61±1.89          & 72.49±1.81          & 54.04±1.10           & 47.40±1.97           & 71.77±0.83          & 55.12±1.03          & 75.92±0.55          & 70.21±0.28          & 76.74±0.82          & 66.01±0.93          \\
\midrule
BERT                             & 37.86±5.31          & 32.78±5.01          & 43.26±1.25           & 34.97±1.58           & 50.14±0.68          & 42.96±1.02          & 54.00±0.20          & 47.57±0.50          & 46.39±1.05          & 37.07±0.68          \\
LLM-GNN                          & 76.15±0.34          & 72.31±1.03          & 63.86±1.85           & 57.16±1.61           & 82.19±0.60          & 75.86±0.30          & 80.78±0.78          & 75.12±0.97          & 78.84±1.29          & 67.15±1.29          \\
GraphTranslator                      & 79.13±1.38          & 74.26±0.93          & 67.55±1.53           & 56.53±1.71           & 80.72±0.80          & 74.20±0.36          & 82.02±0.61          & 75.16±0.48          & 81.01±0.21          & 69.27±1.27          \\
\midrule
G2P2                             & {\ul 80.08±1.33}    & {\ul 75.91±1.39}    & {\ul 68.24±0.53}     & {\ul 58.35±0.35}     & {\ul 82.74±1.98}    & {\ul 76.10±1.59}    & {\ul 82.40±0.90}    & {\ul 76.32±1.04}    & {\ul 81.13±1.06}    & {\ul 69.48±0.15}    \\
\midrule
\method                            & \textbf{82.66±0.77} & \textbf{79.05±1.25} & \textbf{70.79±1.09}  & \textbf{62.72±1.21}  & \textbf{87.99±0.64} & \textbf{82.61±0.81} & \textbf{85.75±0.31} & \textbf{80.45±0.25} & \textbf{85.55±0.58} & \textbf{75.59±0.16} \\
Gain                             & +3.2\%              & +4.1\%              & +3.7\%               & +7.5\%               & +6.3\%              & +8.6\%              & +4.3\%              & +5.4\%              & +5.4\%              & +8.8\%        \\     
\bottomrule

\end{tabular}
}
\caption{Accuracy for few-shot node classification (mean±std). The best and runner-up are marked with bold and underlined, respectively. \textit{Gain} is the relative improvement of \method over the best-performing baseline.}
\label{tab:few-shot classification}
\end{table*}

\begin{table*}[t]
\centering
\resizebox{\textwidth}{!}{
\begin{tabular}{l|cc|cc|cc|cc|cc}
\toprule
\multirow{2}{*}{\textbf{Method}} & \multicolumn{2}{c|}{\textbf{Cora}} & \multicolumn{2}{c|}{\textbf{Fitness}}         & \multicolumn{2}{c|}{\textbf{M.I.}}         & \multicolumn{2}{c|}{\textbf{Industrial}}   & \multicolumn{2}{c}{\textbf{Art}}          \\ \cline{2-11}
                                 & ACC                 & F1                  & ACC                  & F1                   & ACC                 & F1                  & ACC                 & F1                  & ACC                 & F1                  \\
\midrule
BERT                             & 23.56±1.48          & 17.92±0.86          & 32.63±1.24           & 26.58±1.21           & 37.42±0.67          & 30.73±0.93          & 36.88±0.56          & 29.46±1.12          & 35.72±1.59          & 24.10±1.06          \\
LLM-GNN                            & 62.67±1.43          & 55.21±1.47          & 42.47±1.01           & 35.13±1.55           & 70.81±0.83          & 62.66±0.84          & 74.51±1.09          & 63.54±1.56          & 74.63±1.46          & 62.68±1.11          \\
GraphTranslator                          & 62.65±1.85          & 56.92±0.41          & 44.07±1.57           & 38.09±1.89           & 72.67±1.43          & 64.72±0.34          & 74.58±1.31          & 65.13±1.08          & 73.77±0.98          & 61.20±0.65          \\
G2P2                             & {\ul 64.35±2.78}    & {\ul 58.42±1.59}    & {\ul 45.99±0.69}     & {\ul 40.06±1.35}     & {\ul 74.77±1.98}    & {\ul 67.10±1.59}    & {\ul 75.66±1.42}    & {\ul 68.27±1.31}    & {\ul 75.84±1.57}    & {\ul 63.59±1.62}    \\
\midrule
\method                            & \textbf{69.21±1.35} & \textbf{61.41±1.82} & \textbf{54.41±1.10}  & \textbf{47.45±1.63}  & \textbf{79.85±1.35} & \textbf{72.58±0.79} & \textbf{81.99±0.58} & \textbf{73.84±0.33} & \textbf{78.22±1.70} & \textbf{67.71±0.02} \\
Gain                             & +7.6\%              & +5.1\%              & +18.3\%              & +18.4\%              & +6.8\%              & +8.2\%              & +8.4\%              & +8.2\%              & +3.1\%              & +6.5\%          \\   
\bottomrule
\end{tabular}
}
\caption{Accuracy for zero-shot node classification (mean±std). The best and runner-up are marked with bold and underlined, respectively. \textit{Gain} is the relative improvement of \method over the best-performing baseline.}
\label{tab:zero-shot classification}
\vspace{-0.4cm}
\end{table*}

\stitle{Baselines.} We compare \method with 13 baselines from 5 categories, briefly describe as follow. 
\begin{itemize}[leftmargin=*]
    \item \textbf{Supervised GNNs}: GCN~\cite{kipf2016semi}, SAGEsup~\cite{hamilton2017inductive}, TextGCN~\cite{yao2019graph}.  They are trained in a supervised or semi-supervised manner for the node classification tasks.
    \item \textbf{Self-supervised GNNs}: GraphCL~\cite{you2020graph}, InfoGCL~\cite{xu2021infogcl}, PGCL~\cite{lin2022prototypical}. They are first pre-trained via contrastive learning and then fine-tuned for the classification tasks.
    \item \textbf{Graph prompt methods}: GPPT~\cite{sun2022gppt}, GFP~\cite{fang2024universal}, GraphPromt~\cite{liu2023graphprompt}. They reduce the divergence between the pre-training and inference by designing the training objectives and prompts.
    \item \textbf{Language models}: BERT~\cite{devlin2018bert}, LLM-GNN~\cite{chen2023label}, GraphTranslator~\cite{zhang2024graphtranslator}. Bert is first pre-trained and then fine-tuned for text classification. LLM-GNN and GraphTranslator translate graph into language and predict the labels by LLMs.
    \item \textbf{Co-trained model}: G2P2~\cite{wen2023augmenting}. It employs the contrastive loss to train the GNN and language model jointly such that they produce similar node embedding and text embedding for each node-text pair.
\end{itemize}

Following G2P2, we use classification accuracy and F1 score to measure performance. We report the average value and standard deviation across 5 runs. Note that we only select language models and G2P2 as the baselines for zero-shot classification, since the other baselines require at least one labeled sample per class for either training or inference.

\stitle{Environment.} During pre-training, we use Adam as the optimizer with a learning rate of 2e-5 for 2 epochs, and the batch size is 64. The number of similar text representations and the capacity of the text bank are set to 1 and 32K, respectively. The length of the learnable negative prompt is 16. Margin $m$ is set to 1. For few-shot pre-training, we do not activate the negative semantics contrast loss, so $\alpha$ is set to 0. Instead, $\alpha$ is set to 0.5 during zero-shot pre-training. Our experiments are conducted on a server with  Intel(R) Xeon(R) Platinum 8375C CPU @ 2.90GHz and 8 NVIDIA RTX A6000 48G GPUs.

\stitle{Task configurations.} For few-shot classification, we use a 5-way 5-shot setup, i.e., 5 classes are taken from all classes, and then 5 nodes are sampled from these classes to construct the training set. The validation set is generated in the same way as the training set, and all remaining data is used as the test set. For zero-shot classification, we use 5-way classification, which samples classes but does not provide labeled nodes.

\subsection{Main Results (RQ1)}
\stitle{Few-shot node classification.} Table~\ref{tab:few-shot classification} reports the accuracy of \method and the baselines for few-shot node classification. We can see that \method consistently outperforms all baselines across the datasets, with an average improvement of 4.6\% and 6.9\% for classification accuracy and F1 score, respectively. Moreover, the improvements of \method over the baselines are over than 5\% in 6 out of the 10 cases. The results show that \method can mine text semantics effectively to enhance model pre-training and thus improve accuracy. More explanations are in Appendix C.1.

\stitle{Zero-shot node classification.} Table~\ref{tab:zero-shot classification} reports the accuracy of \method and the baselines for zero-shot node classification. We only include the language models and G2P2 because the other methods require at least one labeled sample for inference. The results show that \method consistently outperform all baselines by a large margin. Compared with the best-performing baseline G2P2, the average improvements of \method in classification accuracy and F1 score are  8.8\% and 9.3\%, respectively. All methods have lower accuracy for zero-shot classification than few-shot classification because zero-shot classification does not provided labeled samples, and thus the task is more challenging. However, the improvements of \method are larger for zero-shot classification because it introduces more text semantics for learning.

\stitle{Robustness to task configuration.} We conduct the fewer-way and fewer-shots classification on M.I. dataset. The experimental results are provided in Appendix C.2. The results show that \method outperforms G2P2 across different configurations of ways and shots. We observe that to achieve the same accuracy, \method requires fewer labeled samples (i.e., shots) than G2P2. Moreover, \method surpasses the G2P2 across the different ways for zero-shot classification.

\begin{table}[t]
\centering
\resizebox{\linewidth}{!}{
\begin{tabular}{l|l|ccc}
\toprule
Setting                    & Loss          & M.I.                & Industrial          & Art                 \\
\midrule
\multirow{3}{*}{Few-shot}  & $\mathcal{L}_{CL}$           & 82.74±1.98          & 82.40±0.90          & 81.13±1.06          \\
                           & $\mathcal{L}_{CL+PSM}$      & \textbf{87.91±0.59} & \textbf{85.75±0.31} & \textbf{85.37±0.60} \\
                           & $\mathcal{L}_{CL+PSM+NSC}$ & 87.80±0.28  & 85.63±0.41 & 85.29±0.66 \\
\midrule
\multirow{3}{*}{Zero-shot} & $\mathcal{L}_{CL}$           & 74.77±1.98          & 75.66±1.42          & 75.84±1.57          \\
                           & $\mathcal{L}_{CL+PSM}$      & 78.32±1.22          & 81.85±0.55          & 79.48±1.88          \\
                           & $\mathcal{L}_{CL+PSM+NSC}$ & \textbf{79.15±1.35} & \textbf{81.99±0.58} & \textbf{80.22±1.70} \\
\bottomrule
\end{tabular}
}
\caption{Ablation study of our augmentation techniques. $\mathcal{L}_{CL}$ is the contrastive loss for baseline, \textit{PSM} for positive semantics matching, and \textit{NSC} for negative semantics contrast. Best accuracy in \textbf{bold}.}
\label{tab:loss combination}
\end{table}

\subsection{Micro Experiments}
\label{sec:ablation study}
\stitle{Effect of the augmentations (RQ2).} Table~\ref{tab:loss combination} presents an ablation study by gradually enabling different augmentations techniques in \method. We can see that all augmentations are effective in improving accuracy, as adding each of them outperforms the baseline. The best-performing combination for few-shot classification deactivates negative semantics contrast (i.e., $\mathcal{L}_{NSC}$) while zero-shot classification actives $\mathcal{L}_{NSC}$. This is because few-shot classification uses labeled samples to learn the prompt, and the negative prompt learned by $\mathcal{L}_{NSC}$ may interfere with prompt tuning. In contrast, zero-shot classification lacks labeled data for prompt tuning, and the $\mathcal{L}_{NSC}$ can provide more text semantics.

\begin{figure}[t]
    \centering
    \includegraphics[scale=0.28]{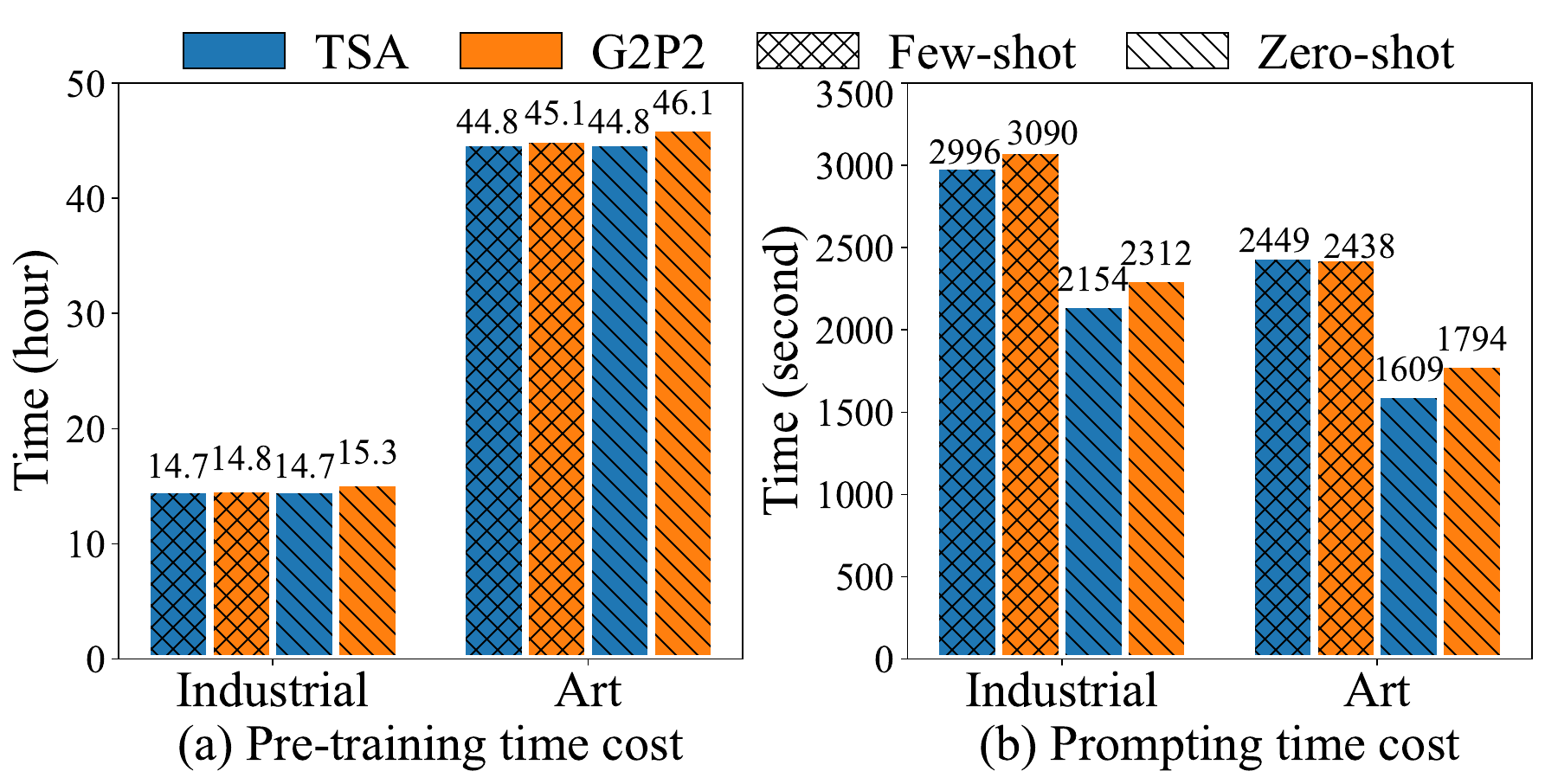}
    \vspace{-0.2cm}
    \caption{The time cost comparison of pre-training and prompting for G2P2 and \method.}
    \label{fig:time}
    \vspace{-0.2cm}
\end{figure}

\stitle{Efficiency (RQ3).} To examine the efficiency of \method, we compare with G2P2 for pre-training time and prompting time at inference time. We experiment on Industrial and Art, the two largest datasets, as the running time is shorter on the smaller datasets. Figure~\ref{fig:time} shows that \method and G2P2 have similar pre-training time and prompting time. This is because they both jointly train the GNN and language model, and computing the loss terms has a small cost compared with computing the two models.
Few-shot classification has longer prompting time than zero-shot classification because it needs to tune the prompt using the labeled samples.

\begin{figure}[t]
    \centering
    \includegraphics[scale=0.28]{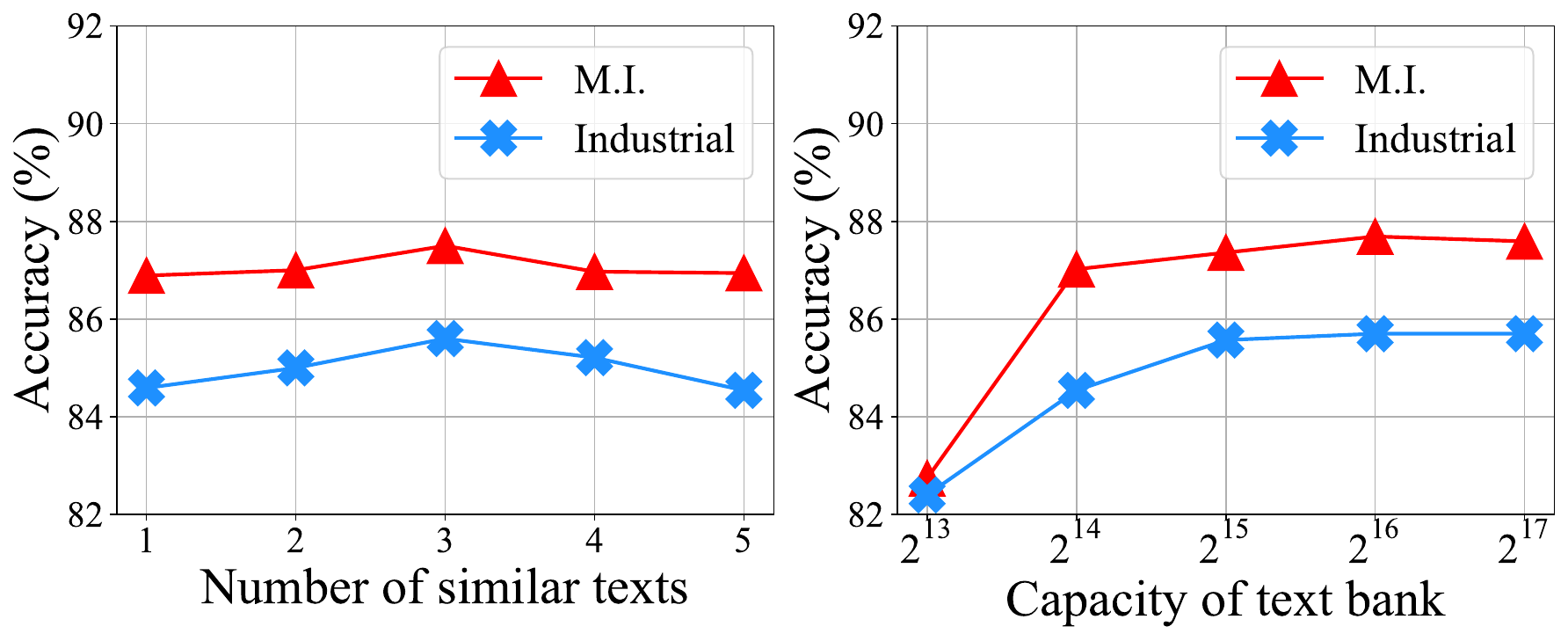}
    \vspace{-0.2cm}
    \caption{The comparison of the number of similar texts and the capacity of text bank for \method on M.I. anf Industrial.}
    \label{fig:text bank}
    \vspace{-0.3cm}
\end{figure}

\stitle{Hyperparameter sensitivity.} We conduct the hyperparameter experiments with respect to the number of similar texts and the capacity of text bank. Figure~\ref{fig:text bank} examines the effect of the two parameters on the Industrial and M.I. datasets. We observe that accuracy first increases but then decreases with the number of similar texts. This is because while more similar texts can provide more text semantics, an excessive number of these signals may introduce noise by including texts that are not truly similar to the target node. 
Hence, the optimal accuracy are obtained at an intermediate value to balance between semantic supervisions and noises. 
When increasing the capacity of the text bank, accuracy first increases but then stabilizes. This is because using a larger text bank allows a node to identify texts that are more similar but the similarity will become sufficiently highly when the bank is large enough.

\section{Related Work}
\stitle{Graph Pre-training and Prompting.}  GNNs~\cite{kipf2016semi,velivckovic2017graph} use message passing to aggregate features from neighboring nodes to compute graph node embedding. However, early GNN models, such as  GCN~\cite{kipf2016semi}, and GAT~\cite{velivckovic2017graph}, are supervised and require many labeled nodes for training. To mine supervision signals from unlabeled data, graph self-supervised learning is proposed to train using well-designed pretext tasks. For instance, DGI~\cite{velivckovic2018deep} learns node embeddings by maximizing mutual information between the global and local node embeddings.  GPT-GNN~\cite{hu2020gpt} utilizes a self-supervised graph generation task to combine the graph structural and semantic information.

Graph self-supervised learning~\cite{wang2024gcmae,wang2024sgdc,LiangZS0X0T025,wang2025tesa,ZhangZZGWLXZJQZY25} methods still require many labeled to fine-tune specific tasks (e.g., node classification). To further reduce the reliance on labeled instances, graph prompt learning is proposed for few-shot node classification. For example, GPPT predicts the node label by deciding whether an edge exists between the target node and candidate labels. GraphPrompt~\cite{liu2023graphprompt} learns embeddings for subgraphs rather than nodes to unify graph-level and node-level tasks. These approaches consider only the graph and thus have limited accuracy for TAGs with text descriptions. To account for the text, TextGCN~\cite{yao2019graph} generates text embeddings using pre-trained language models and adds these embeddings as node features for GNN training. G2P2~\cite{wen2023augmenting} jointly trains the language model and GNN with the contrastive strategy and uses prompting for few-shot and zero-shot node classification. 
However,  \method targets TAGs and considers the graph and text modalities jointly by mining more text semantics while graph pre-training methods consider only the graph.

\stitle{Pre-trained Language Models (PLMs).} PLMs~\cite{devlin2018bert,yang2022ocenbase,yang2023ocenbase,han2024palf} enhance the ability to understand natural language by pre-training on large-scale text corpus. The well-known BERT~\cite{devlin2018bert}, for instance, is pre-trained with two tasks, i.e., masked token reconstruction and next token prediction, to capture contextual information. RoBERTa~\cite{liu2019roberta} improves BERT by eliminating the next token prediction task, increasing the batch size and data volume during pre-training, and using a dynamic masking strategy. While PLMs achieve great success for text oriented tasks, they cannot capture the topology information for TAGs.

\section{Conclusion}
In this paper, we study few-shot and zero-shot node classification on text-attributed graphs. We observe that the prior methods is limited to graph-based augmentation techniques, thus we propose \method as a novel pre-training and inference framework. \method incorporates two key augmentation techniques, i.e., positive semantics matching and negative semantics contrast, to exploit more text semantics. Extensive experiments show that \method outperforms existing methods by a large margin. 
We believe our methodology, i.e., generating node-text pairs that have similar/dissimilar embeddings, is general and can be extended beyond our augmentation techniques. 

\clearpage

\appendix
\section{Proof of Theorem 1}
We prove Theorem 1 in detail as follows:

\begin{proof}
$H(\mathbf{X}^{neg})$ is equal to the sum of the information entropy of the negation words and that of the raw text as follow:
    \begin{equation}
    \resizebox{0.91\linewidth}{!}{$
        H(\mathbf{X}^{neg})= \!-\! {\textstyle \sum_{i}}P(\mathbf{x}^{neg}_{i})logP(\mathbf{x}^{neg}_{i}) \!-\!{\textstyle \sum_{j\ne i}P(\mathbf{x}_{j})logP(\mathbf{x}_{j})}.
        $}
    \end{equation}
    Similarly, $H(\mathbf{h})$ can be calculated by the following equation:
    \begin{equation}
        \resizebox{0.91\linewidth}{!}{$
        H(\mathbf{h})=-{\textstyle \sum_{i}^{M}}P(\mathrm{V}_{i})logP(\mathrm{V}_{i}) \!-\! {\textstyle \sum_{j}}P(\mathbf{x}_{j})logP(\mathbf{x}_{j}).
        $}
        \label{eq:entropy}
    \end{equation}
    Due to the concentration of the probability distribution of characters on the negation words, the lower bound of the $H(\mathbf{X}^{neg})$ can be approximately equal to:
    \begin{equation}
        LowerBound(H(\mathbf{X}^{neg})) \!\approx \!-\! {\textstyle \sum_{i}}P(x^{neg}_{i})logP(x^{neg}_{i}) 
    \end{equation}
    Since the negative prompts $\mathrm{V}$ are the high-dimensional vectors, their probability distributions are more dispersed. According to the Maximum Entropy Theorem, the more dispersed the distribution, the greater the entropy:
    \begin{equation}
        -{\textstyle \sum_{i}^{M}}P(\mathrm{V}_{i})logP(\mathrm{V}_{i}) \ge LowerBound(H(\mathbf{X}^{neg})).
    \end{equation}
Therefore, similar to Equation~\ref{eq:entropy}, by ignoring the entropy of the raw text, i.e., $- {\textstyle \sum_{j}}P(\mathbf{x}_{j})logP(\mathbf{x}_{j})\!\approx\! 0$ , we have:
\begin{equation}
    H(\mathbf{h})\ge {\tt LowerBound}(H(\mathbf{X}^{neg})).
\end{equation}
\end{proof}

\section{Dataset Statistics}

We use five public datasets for experimental evaluation. See Table~\ref{tab:dataset statistic} for the dataset statistics.

\begin{table}[h]
\resizebox{\linewidth}{!}{
\begin{tabular}{l|rrrr}
\bottomrule
\textbf{Dataset} & \textbf{\# Nodes} & \textbf{\# Edges} & \textbf{\# Avg.deg} & \textbf{\# Classes} \\
\midrule
Cora             & 25,120           & 182,280          & 7.26                 & 70                 \\
Fitness          & 173,055          & 1,773,500        & 17.45                 & 13   \\
M.I.             & 905,453          & 2,692,734        & 2.97                 & 1,191              \\
Industrial       & 1,260,053        & 3,101,670        & 2.46                 & 2,462              \\
Art              & 1,615,902        & 4,898,218        & 3.03                 & 3,347              \\
\bottomrule
\end{tabular}
}
\caption{Statistics of the experiment datasets.}
\label{tab:dataset statistic}
\end{table}

\section{Experimental Result and Analysis}
\subsection{Analysis for few-shot classification}
We provide more experimental analysis for the few-shot node classification.
On Cora dataset, the improvements of \method are smaller than the other datasets because Cora is the smallest among the datasets, and thus existing methods learn relatively well.
Regarding the baselines, supervised GNNs have the lowest accuracy since they are only a few labeled nodes for predicting. Contrastive GNNs outperform supervised GNNs, suggesting that self-supervised pre-training is important.
Graph prompt methods only utilize the graph structures and neglect the text descriptions. Conversely, language models only use the text descriptions and ignore the graph structures. G2P2 jointly trains the GNN and language model, and thus it achieves the best performance among the baselines. Nonetheless, \method outperforms G2P2 because it exploits more text semantics with our two augmentations.

\begin{figure}[t]
    \centering
    \includegraphics[scale=0.28]{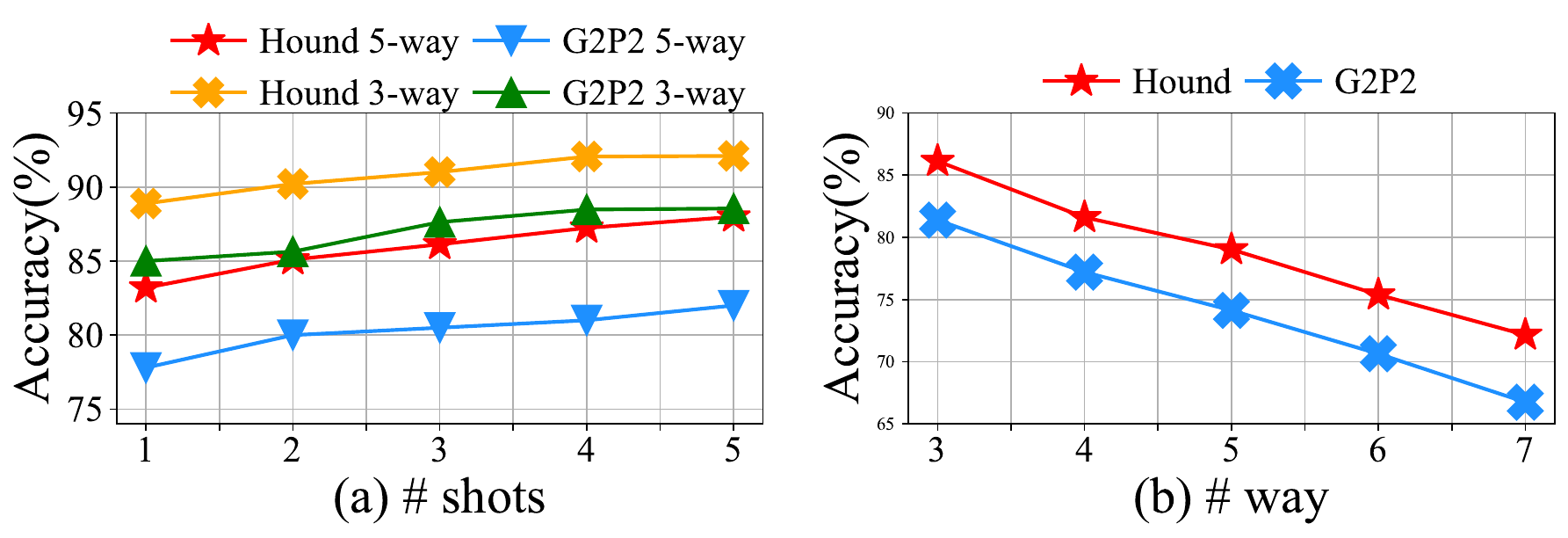}
    \caption{The accuracy comparison for our \method and G2P2 in the fewer-way and fewer-shot settings on M.I. dataset.}
    \label{fig:way-shot}
\end{figure}

\subsection{Robustness to task configuration}
In Figure~\ref{fig:way-shot}, we experiment with 3-way and 5-way classification and change the number of shots (i.e., number of labeled samples from each class) for few-shot classification. There are not labeled samples for zero-shot classification, thus we only change the number of ways. The results show that \method surpasses  G2P2 on all different configurations. 
Both \method and G2P2 perform better for 3-way classification than 5-way classification because 3-way classification is easier. Moreover, accuracy improves with the number of shots because there are more labeled samples. The accuracy of both \method and G2P2 decreases with the number of way as the difficulty of classification increases.



\bibliographystyle{named}
\bibliography{main}

\begin{thebibliography}{}

\bibitem[\protect\citeauthoryear{Chen \bgroup \em et al.\egroup }{2023}]{chen2023label}
Zhikai Chen, Haitao Mao, Hongzhi Wen, Haoyu Han, Wei Jin, Haiyang Zhang, Hui Liu, and Jiliang Tang.
\newblock Label-free node classification on graphs with large language models (llms).
\newblock {\em arXiv preprint arXiv:2310.04668}, 2023.

\bibitem[\protect\citeauthoryear{Chen \bgroup \em et al.\egroup }{2024}]{chen2024exploring}
Zhikai Chen, Haitao Mao, Hang Li, Wei Jin, Hongzhi Wen, Xiaochi Wei, Shuaiqiang Wang, Dawei Yin, Wenqi Fan, Hui Liu, et~al.
\newblock Exploring the potential of large language models (llms) in learning on graphs.
\newblock {\em ACM SIGKDD Explorations Newsletter}, 25(2):42--61, 2024.

\bibitem[\protect\citeauthoryear{Chuang \bgroup \em et al.\egroup }{2020}]{chuang2020debiased}
Ching-Yao Chuang, Joshua Robinson, Yen-Chen Lin, Antonio Torralba, and Stefanie Jegelka.
\newblock Debiased contrastive learning.
\newblock {\em Advances in Neural Information Processing Systems}, 33:8765--8775, 2020.

\bibitem[\protect\citeauthoryear{Deng and Hooi}{2021}]{deng2021graph}
Ailin Deng and Bryan Hooi.
\newblock Graph neural network-based anomaly detection in multivariate time series.
\newblock In {\em Proceedings of the AAAI Conference on Artificial Intelligence}, volume~35, pages 4027--4035, 2021.

\bibitem[\protect\citeauthoryear{Devlin \bgroup \em et al.\egroup }{2018}]{devlin2018bert}
Jacob Devlin, Ming-Wei Chang, Kenton Lee, and Kristina Toutanova.
\newblock Bert: pre-training of deep bidirectional transformers for language understanding.
\newblock {\em arXiv preprint arXiv:1810.04805}, 2018.

\bibitem[\protect\citeauthoryear{Fang \bgroup \em et al.\egroup }{2024}]{fang2024universal}
Taoran Fang, Yunchao Zhang, Yang Yang, Chunping Wang, and Lei Chen.
\newblock Universal prompt tuning for graph neural networks.
\newblock {\em Advances in Neural Information Processing Systems}, 36, 2024.

\bibitem[\protect\citeauthoryear{Gao \bgroup \em et al.\egroup }{2022}]{gao2022graph}
Chen Gao, Xiang Wang, Xiangnan He, and Yong Li.
\newblock Graph neural networks for recommender system.
\newblock In {\em Proceedings of the 15th ACM International Conference on Web Search and Data Mining}, pages 1623--1625, 2022.

\bibitem[\protect\citeauthoryear{Hamilton \bgroup \em et al.\egroup }{2017}]{hamilton2017inductive}
Will Hamilton, Zhitao Ying, and Jure Leskovec.
\newblock Inductive representation learning on large graphs.
\newblock {\em Advances in Neural Information Processing Systems}, 30, 2017.

\bibitem[\protect\citeauthoryear{Han \bgroup \em et al.\egroup }{2024}]{han2024palf}
Fusheng Han, Hao Liu, Bin Chen, Debin Jia, Jianfeng Zhou, Xuwang Teng, Chuanhui Yang, Huafeng Xi, Wei Tian, Shuning Tao, Sen Wang, Quanqing Xu, and Zhenkun Yang.
\newblock Palf: Replicated write-ahead logging for distributed databases.
\newblock {\em Proc. VLDB Endow.}, 17(12):3745–3758, August 2024.

\bibitem[\protect\citeauthoryear{He \bgroup \em et al.\egroup }{2020}]{he2020momentum}
Kaiming He, Haoqi Fan, Yuxin Wu, Saining Xie, and Ross Girshick.
\newblock Momentum contrast for unsupervised visual representation learning.
\newblock In {\em Proceedings of the IEEE/CVF Conference on Computer Vision and Pattern Recognition}, pages 9729--9738, 2020.

\bibitem[\protect\citeauthoryear{Hu \bgroup \em et al.\egroup }{2020}]{hu2020gpt}
Ziniu Hu, Yuxiao Dong, Kuansan Wang, Kai-Wei Chang, and Yizhou Sun.
\newblock Gpt-gnn: Generative pre-training of graph neural networks.
\newblock In {\em Proceedings of the 26th ACM SIGKDD International Conference on Knowledge Discovery and Data Mining}, pages 1857--1867, 2020.

\bibitem[\protect\citeauthoryear{Huang \bgroup \em et al.\egroup }{2023}]{huang2023prompt}
Xuanwen Huang, Kaiqiao Han, Dezheng Bao, Quanjin Tao, Zhisheng Zhang, Yang Yang, and Qi~Zhu.
\newblock Prompt-based node feature extractor for few-shot learning on text-attributed graphs.
\newblock {\em arXiv preprint arXiv:2309.02848}, 2023.

\bibitem[\protect\citeauthoryear{Kipf and Welling}{2016}]{kipf2016semi}
Thomas~N Kipf and Max Welling.
\newblock Semi-supervised classification with graph convolutional networks.
\newblock {\em arXiv preprint arXiv:1609.02907}, 2016.

\bibitem[\protect\citeauthoryear{Liang \bgroup \em et al.\egroup }{2025}]{LiangZS0X0T025}
Yuxuan Liang, Wentao Zhang, Zeang Sheng, Ling Yang, Quanqing Xu, Jiawei Jiang, Yunhai Tong, and Bin Cui.
\newblock Towards scalable and deep graph neural networks via noise masking.
\newblock In {\em Sponsored by the Association for the Advancement of Artificial Intelligence}, pages 18693--18701, 2025.

\bibitem[\protect\citeauthoryear{Lin \bgroup \em et al.\egroup }{2022}]{lin2022prototypical}
Shuai Lin, Chen Liu, Pan Zhou, Zi-Yuan Hu, Shuojia Wang, Ruihui Zhao, Yefeng Zheng, Liang Lin, Eric Xing, and Xiaodan Liang.
\newblock Prototypical graph contrastive learning.
\newblock {\em IEEE Transactions on Neural Networks and Learning Systems}, 35(2):2747--2758, 2022.

\bibitem[\protect\citeauthoryear{Liu \bgroup \em et al.\egroup }{2019}]{liu2019roberta}
Yinhan Liu, Myle Ott, Naman Goyal, Jingfei Du, Mandar Joshi, Danqi Chen, Omer Levy, Mike Lewis, Luke Zettlemoyer, and Veselin Stoyanov.
\newblock Roberta: a robustly optimized bert pretraining approach.
\newblock {\em arXiv preprint arXiv:1907.11692}, 2019.

\bibitem[\protect\citeauthoryear{Liu \bgroup \em et al.\egroup }{2021}]{liu2021relative}
Zemin Liu, Yuan Fang, Chenghao Liu, and Steven~CH Hoi.
\newblock Relative and absolute location embedding for few-shot node classification on graph.
\newblock In {\em Proceedings of the AAAI Conference on Artificial Intelligence}, volume~35, pages 4267--4275, 2021.

\bibitem[\protect\citeauthoryear{Liu \bgroup \em et al.\egroup }{2022}]{liu2022few}
Yonghao Liu, Mengyu Li, Ximing Li, Fausto Giunchiglia, Xiaoyue Feng, and Renchu Guan.
\newblock Few-shot node classification on attributed networks with graph meta-learning.
\newblock In {\em Proceedings of the 45th international ACM SIGIR Conference on Research and Development in Information Retrieval}, pages 471--481, 2022.

\bibitem[\protect\citeauthoryear{Liu \bgroup \em et al.\egroup }{2023}]{liu2023graphprompt}
Zemin Liu, Xingtong Yu, Yuan Fang, and Xinming Zhang.
\newblock Graphprompt: Unifying pre-training and downstream tasks for graph neural networks.
\newblock In {\em Proceedings of the ACM Web Conference 2023}, pages 417--428, 2023.

\bibitem[\protect\citeauthoryear{McCallum \bgroup \em et al.\egroup }{2000}]{mccallum2000automating}
Andrew~Kachites McCallum, Kamal Nigam, Jason Rennie, and Kristie Seymore.
\newblock Automating the construction of internet portals with machine learning.
\newblock {\em Information Retrieval}, 3:127--163, 2000.

\bibitem[\protect\citeauthoryear{Noble and Cook}{2003}]{noble2003graph}
Caleb~C Noble and Diane~J Cook.
\newblock Graph-based anomaly detection.
\newblock In {\em Proceedings of the ninth ACM SIGKDD International Conference on Knowledge Discovery and Data Mining}, pages 631--636, 2003.

\bibitem[\protect\citeauthoryear{Sun \bgroup \em et al.\egroup }{2022}]{sun2022gppt}
Mingchen Sun, Kaixiong Zhou, Xin He, Ying Wang, and Xin Wang.
\newblock Gppt: Graph pre-training and prompt tuning to generalize graph neural networks.
\newblock In {\em Proceedings of the 28th ACM SIGKDD Conference on Knowledge Discovery and Data Mining}, pages 1717--1727, 2022.

\bibitem[\protect\citeauthoryear{Tang \bgroup \em et al.\egroup }{2024}]{tang2024graphgpt}
Jiabin Tang, Yuhao Yang, Wei Wei, Lei Shi, Lixin Su, Suqi Cheng, Dawei Yin, and Chao Huang.
\newblock Graphgpt: Graph instruction tuning for large language models.
\newblock In {\em Proceedings of the 47th International ACM SIGIR Conference on Research and Development in Information Retrieval}, pages 491--500, 2024.

\bibitem[\protect\citeauthoryear{Vaswani \bgroup \em et al.\egroup }{2017}]{vaswani2017attention}
Ashish Vaswani, Noam Shazeer, Niki Parmar, Jakob Uszkoreit, Llion Jones, Aidan~N Gomez, {\L}ukasz Kaiser, and Illia Polosukhin.
\newblock Attention is all you need.
\newblock {\em Advances in Neural Information Processing Systems}, 30, 2017.

\bibitem[\protect\citeauthoryear{Veli{\v{c}}kovi{\'c} \bgroup \em et al.\egroup }{2017}]{velivckovic2017graph}
Petar Veli{\v{c}}kovi{\'c}, Guillem Cucurull, Arantxa Casanova, Adriana Romero, Pietro Lio, and Yoshua Bengio.
\newblock Graph attention networks.
\newblock {\em arXiv preprint arXiv:1710.10903}, 2017.

\bibitem[\protect\citeauthoryear{Veli{\v{c}}kovi{\'c} \bgroup \em et al.\egroup }{2018}]{velivckovic2018deep}
Petar Veli{\v{c}}kovi{\'c}, William Fedus, William~L Hamilton, Pietro Li{\`o}, Yoshua Bengio, and R~Devon Hjelm.
\newblock Deep graph infomax.
\newblock {\em arXiv preprint arXiv:1809.10341}, 2018.

\bibitem[\protect\citeauthoryear{Wang \bgroup \em et al.\egroup }{2024a}]{wang2024gcmae}
Yuxiang Wang, Xiao Yan, Chuang Hu, Quanqing Xu, Chuanhui Yang, Fangcheng Fu, Wentao Zhang, Hao Wang, Bo~Du, and Jiawei Jiang.
\newblock Generative and contrastive paradigms are complementary for graph self-supervised learning.
\newblock In {\em 2024 IEEE 40th International Conference on Data Engineering}, pages 3364--3378, 2024.

\bibitem[\protect\citeauthoryear{Wang \bgroup \em et al.\egroup }{2024b}]{wang2024sgdc}
Yuxiang Wang, Xiao Yan, Shiyu Jin, Hao Huang, Quanqing Xu, Qingchen Zhang, Bo~Du, and Jiawei Jiang.
\newblock Self-supervised learning for graph dataset condensation.
\newblock In {\em Proceedings of the 30th ACM SIGKDD Conference on Knowledge Discovery and Data Mining}, page 3289–3298. Association for Computing Machinery, 2024.

\bibitem[\protect\citeauthoryear{Wang \bgroup \em et al.\egroup }{2025}]{wang2025tesa}
Xin Wang, Jiawei Jiang, Xiao Yan, and Qiang Huang.
\newblock {TESA:} {A} trajectory and semantic-aware dynamic heterogeneous graph neural network.
\newblock In {\em Proceedings of the ACM on Web Conference}, pages 1305--1315. {ACM}, 2025.

\bibitem[\protect\citeauthoryear{Wen and Fang}{2023}]{wen2023augmenting}
Zhihao Wen and Yuan Fang.
\newblock Augmenting low-resource text classification with graph-grounded pre-training and prompting.
\newblock In {\em Proceedings of the 46th International ACM SIGIR Conference on Research and Development in Information Retrieval}, pages 506--516, 2023.

\bibitem[\protect\citeauthoryear{Xu \bgroup \em et al.\egroup }{2021}]{xu2021infogcl}
Dongkuan Xu, Wei Cheng, Dongsheng Luo, Haifeng Chen, and Xiang Zhang.
\newblock Infogcl: Information-aware graph contrastive learning.
\newblock {\em Advances in Neural Information Processing Systems}, 34:30414--30425, 2021.

\bibitem[\protect\citeauthoryear{Yan \bgroup \em et al.\egroup }{2023}]{yan2023comprehensive}
Hao Yan, Chaozhuo Li, Ruosong Long, Chao Yan, Jianan Zhao, Wenwen Zhuang, Jun Yin, Peiyan Zhang, Weihao Han, Hao Sun, et~al.
\newblock A comprehensive study on text-attributed graphs: Benchmarking and rethinking.
\newblock {\em Advances in Neural Information Processing Systems}, 36:17238--17264, 2023.

\bibitem[\protect\citeauthoryear{Yang \bgroup \em et al.\egroup }{2022}]{yang2022ocenbase}
Zhenkun Yang, Chuanhui Yang, Fusheng Han, Mingqiang Zhuang, Bing Yang, Zhifeng Yang, Xiaojun Cheng, Yuzhong Zhao, Wenhui Shi, Huafeng Xi, Huang Yu, Bin Liu, Yi~Pan, Boxue Yin, Junquan Chen, and Quanqing Xu.
\newblock Oceanbase: a 707 million tpmc distributed relational database system.
\newblock {\em Proc. VLDB Endow.}, 15(12):3385–3397, August 2022.

\bibitem[\protect\citeauthoryear{Yang \bgroup \em et al.\egroup }{2023}]{yang2023ocenbase}
Zhifeng Yang, Quanqing Xu, Shanyan Gao, Chuanhui Yang, Guoping Wang, Yuzhong Zhao, Fanyu Kong, Hao Liu, Wanhong Wang, and Jinliang Xiao.
\newblock Oceanbase paetica: A hybrid shared-nothing/shared-everything database for supporting single machine and distributed cluster.
\newblock {\em Proc. VLDB Endow.}, 16(12):3728–3740, August 2023.

\bibitem[\protect\citeauthoryear{Yao \bgroup \em et al.\egroup }{2019}]{yao2019graph}
Liang Yao, Chengsheng Mao, and Yuan Luo.
\newblock Graph convolutional networks for text classification.
\newblock In {\em Proceedings of the AAAI Conference on Artificial Intelligence}, volume~33, pages 7370--7377, 2019.

\bibitem[\protect\citeauthoryear{You \bgroup \em et al.\egroup }{2020}]{you2020graph}
Yuning You, Tianlong Chen, Yongduo Sui, Ting Chen, Zhangyang Wang, and Yang Shen.
\newblock Graph contrastive learning with augmentations.
\newblock {\em Advances in Neural Information Processing Systems}, 33:5812--5823, 2020.

\bibitem[\protect\citeauthoryear{Yu \bgroup \em et al.\egroup }{2020}]{yu2020enhancing}
Junliang Yu, Hongzhi Yin, Jundong Li, Min Gao, Zi~Huang, and Lizhen Cui.
\newblock Enhancing social recommendation with adversarial graph convolutional networks.
\newblock {\em IEEE Transactions on Knowledge and Data Engineering}, 34(8):3727--3739, 2020.

\bibitem[\protect\citeauthoryear{Zhang \bgroup \em et al.\egroup }{2024}]{zhang2024graphtranslator}
Mengmei Zhang, Mingwei Sun, Peng Wang, Shen Fan, Yanhu Mo, Xiaoxiao Xu, Hong Liu, Cheng Yang, and Chuan Shi.
\newblock Graphtranslator: Aligning graph model to large language model for open-ended tasks.
\newblock In {\em Proceedings of the ACM on Web Conference 2024}, pages 1003--1014, 2024.

\bibitem[\protect\citeauthoryear{Zhang \bgroup \em et al.\egroup }{2025}]{ZhangZZGWLXZJQZY25}
Yongliang Zhang, Yuanyuan Zhu, Hao Zhang, Congli Gao, Yuyang Wang, Guojing Li, Tianyang Xu, Ming Zhong, Jiawei Jiang, Tieyun Qian, Chenyi Zhang, and Jeffrey~Xu Yu.
\newblock Tgraph: A tensor-centric graph processing framework.
\newblock {\em Proc. {ACM} Manag. Data}, 3(1):81:1--81:27, 2025.

\bibitem[\protect\citeauthoryear{Zhao \bgroup \em et al.\egroup }{2024}]{zhao2024pre}
Huanjing Zhao, Beining Yang, Yukuo Cen, Junyu Ren, Chenhui Zhang, Yuxiao Dong, Evgeny Kharlamov, Shu Zhao, and Jie Tang.
\newblock Pre-training and prompting for few-shot node classification on text-attributed graphs.
\newblock {\em arXiv preprint arXiv:2407.15431}, 2024.

\bibitem[\protect\citeauthoryear{Zhou \bgroup \em et al.\egroup }{2022}]{zhou2022conditional}
Kaiyang Zhou, Jingkang Yang, Chen~Change Loy, and Ziwei Liu.
\newblock Conditional prompt learning for vision-language models.
\newblock In {\em Proceedings of the IEEE/CVF Conference on Computer Vision and Pattern Recognition}, pages 16816--16825, 2022.

\end{thebibliography}

\end{document}